\pdfoutput=1
\documentclass[11pt]{article}

\usepackage{ACL2023}

\usepackage{times}
\usepackage{latexsym}
\usepackage{booktabs}
\usepackage{graphicx}
\usepackage{color}

\usepackage[T1]{fontenc}
\usepackage[utf8]{inputenc}

\usepackage{microtype}

\usepackage{inconsolata}

\title{GOAL: A Challenging Knowledge-grounded Video Captioning Benchmark for Real-time Soccer Commentary Generation}

\author{Ji Qi$^{1*}$, Jifan Yu$^{1}\thanks{\quad indicates equal contribution}$, Teng Tu$^1$, Kunyu Gao$^1$, Yifan Xu$^{1}$ Xinyu Guan$^2$, Xiaozhi Wang$^1$,
  \\ \textbf{Yuxiao Dong$^1$, Bin Xu$^{1}$\thanks{\quad Corresponding author: xubin@tsinghua.edu.cn}, Lei Hou$^1$, Juanzi Li$^1$, Jie Tang$^1$, Weidong Guo$^3$, Hui Liu$^3$, Yu Xu$^3$} \\
  $^1$Department of Computer Science and Technology, Tsinghua University, Beijing, China \\
  $^2$Biendata, $^3$Platform and Content Group, Tencent \\
  \texttt{qj20@mails.tsinghua.edu.cn}, \texttt{yujf18@mails.tsinghua.edu.cn}
}

\begin{document}
\maketitle
\begin{abstract}

Despite the recent emergence of video captioning models, how to generate vivid, fine-grained video descriptions based on the background knowledge (\emph{i.e.}, long and informative commentary about the domain-specific scenes with appropriate reasoning) is still far from being solved, which however has great applications such as automatic sports narrative. In this paper, we present GOAL, a benchmark of over $8.9$k soccer video clips, $22$k sentences, and $42$k knowledge triples for proposing a challenging new task setting as Knowledge-grounded Video Captioning (KGVC). Moreover, we conduct
experimental adaption of existing methods to show the difficulty and potential directions for solving this valuable and applicable task.
Our data and code are available at \url{https://github.com/THU-KEG/goal}.

\end{abstract}

\section{Introduction}
\label{sec:introduction}

Video captioning, the task of describing a video's content in natural language~\cite{liu2021video}, was initially proposed due to its importance in a wide range of applications~\cite{venugopalan2015sequence}. Since then, it has been a crucial and challenging task in both computer vision and natural language processing communities, as it requires the mastery of several skills for one model (or a system)~\cite{lin2022swinbert}, including (1) \emph{video understanding}: understanding of the spatial-temporal dynamics in video~\cite{xu2021videoclip}, (2) \emph{video-text bridging}: associating visual and textual elements~\cite{buch2022revisiting,ge2022bridging} and (3) \emph{text generation}: generating appropriate long sequences of output words~\cite{shi2022learning}.

\begin{figure*}[t]
    \centering
    \includegraphics[width=1.0\linewidth]{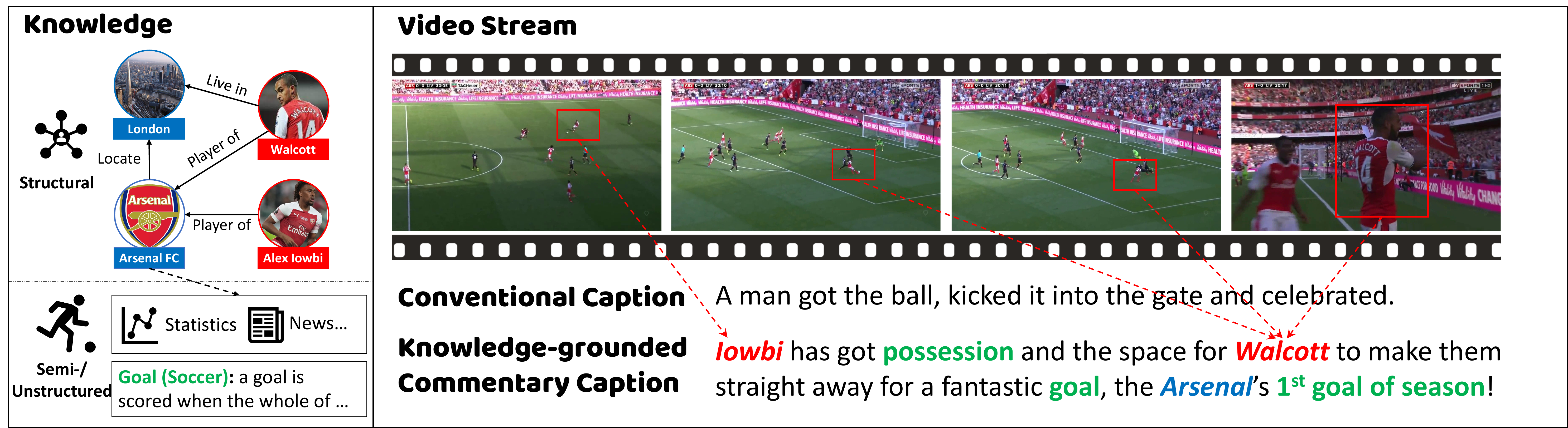}
    \caption{Comparison of conventional video captioning with its knowledge-grounded setting. A soccer commentary describe the fine-grained entities ({\color{red}red}), their relevant structural ({\color{blue}blue}), and semi-/unstructured knowledge ({\color{green}green}).}
    \label{data_example}
\end{figure*}

Despite the recent models' impressive performance~\cite{seo2022end,lin2022swinbert}, it is noteworthy that the existing video captioning benchmarks~\cite{caba2015activitynet,xu2016msr,zhou2018towards}  are still several steps away from the real-world applications. As the soccer narrative examples in Figure \ref{data_example}, when commentators describe the scenes in a video, they need to recognize the visible objects (\emph{Walcott}) and actions (\emph{Goal}), and exploit the relevant knowledge \emph{(Walcott, Player of, Arsenal)} to understand and complete an expression. Due to the involvement of knowledge beyond the video, such applications pose more challenges to all three skills of the video captioning model than the setting of relevant datasets: First, video understanding is additionally required to be able to link the objects to fine-grained entities (\emph{Man}$ \rightarrow$ \emph{Walcott}) and combine the multiple actions to reason certain events (\emph{Kick, Celebrate} $\rightarrow$ \emph{Goal}). Second, besides bridging the video and text, associating the knowledge behind them also becomes a crucial issue. Finally, models need to invoke background knowledge (\emph{$1^{st}$ Goal of season}) to generate vivid descriptions and comments instead of simply introducing the coarse-grained scene.

In this paper, we propose a knowledge-\textbf{G}rounded vide\textbf{O} c\textbf{A}ptioning benchmark for rea\textbf{L}-time soccer commentary generation (\textbf{GOAL}), which provides a more challenging setting of this task. Built upon the collection of over $40$ hours of broadcast soccer videos~\cite{Deliege2020SoccerNetv2}, our dataset contains more than $8.9$k video clips with corresponding commentary texts. Each video in our dataset is linked to abundant knowledge from a professional sports platform\footnote{\url{https://www.whoscored.com/}\label{whoscores}} and the Wikidata~\cite{vrandevcic2014wikidata}. After that, we conduct a series of human annotation tasks, including proofreading, entity recognition, and text classification, to complete the fine-grained alignment of the knowledge, video, and text. Furthermore, we carefully pre-process the videos via object detection, event spot, and 2D/3D feature modeling to provide an easy-to-adapt interface for the state-of-the-art video captioning models.

Experimental results show that the performance of several top-performing video captioning models~\cite{lin2022swinbert,li2022align} suffers an apparent decline on our benchmark, which indicates the difficulty of such a setting. Moreover, we conduct some primary explorations about knowledge-aware video captioning methods. We hope our work can call for more efforts to exploit the advanced NLP techniques (such as large-scale language model (LLM) instructions~\cite{brown2020language,ouyang2022training}) in promoting this task into complex real-world applications.

Our contributions include: a) the proposal of a ``knowledge-grounded'' setting for enriching the video captioning task; b) a high-quality, knowledge-intensive dataset from soccer commentary as a challenging benchmark; c) an investigation on the current state and
the direction for building knowledge-grounded video captioning models.

\section{GOAL Benchmark}

\textbf{Problem Formulation}. Consider a video $V$ consisting of $N_{v}$ frames described by a textual sequence $S$. Distinct from conventional video captioning setting that aims at generating $S_v$ only based on $V$, the task of \textbf{Knowledge-grounded Video Captioning} \textbf{(KGVC)} can be formulated as: given the video $V$, the objective is to select appropriate video-related knowledge $K_{v} = \mathbf{f}(V)$ and generate a knowledgeable description $S_{v,k}$.
\begin{equation}
    \textbf{KGVC}: S_{v,k} = \mathbf{g}(V,K_v) = \mathbf{g}(V,\mathbf{f}(V))
\end{equation}
where functions $\mathbf{f}(\cdot)$ and $\mathbf{g}(\cdot)$ respectively denote the models' abilities on aligning video and knowledge (upon \emph{video understanding}) and converting video and knowledge to knowledgeable texts (upon \emph{video-text bridging} and \emph{text generation}). Therefore, towards a high-quality KGVC benchmark, not only the accuracy of ``Video-Text-Knowledge'' triple is determined to be guaranteed (for training and testing $\mathbf{g}$), but the candidate knowledge of the video is also suggested to be sufficient (for training $\mathbf{f}$ ).

\subsection{Data Collection and Pre-processing}

Broadcast soccer videos with commentary are the foundation of our dataset.To begin the construction of the GOAL benchmark, we collect $80$ full-game videos narrated in English among $500$ games of the open-source SoccerNet-v2~\cite{Deliege2020SoccerNetv2}. Then we employ the Azure ASR toolkit~\footnote{\url{https://azure.microsoft.com/}} to convert the speeches to raw texts. After filtering out videos with low resolution or too sparse narration, only $20$ games are preserved as annotation candidates. Furthermore, we record the basic game information of each video, such as the \emph{gameID}, \emph{matching teams}, \emph{league}, \emph{season}, and \emph{game result} for subsequent information seeking.

\subsection{Video-Text-Knowledge Annotation}

To guarantee the quality of the dataset, we host a team of $10$ English native speakers (and are veteran soccer fans) for a series of human annotation tasks, including 1) \textbf{commentary text proofreading}: as the raw texts are not accurate enough, the first task is to do proofreading of all the texts according to the accompanying video and speech; 2) \textbf{video-text alignment}: along with proofreading, annotators are also required to adjust the text breaks and align them with the video timeline; 3) \textbf{knowledge annotation of text}: for each proofread sentence, annotators recognize the mentions of knowledge entities (especially \emph{players}, \emph{teams} and \emph{soccer terms}) with the help of the given game information and Internet. They also classify each sentence as a scene description, background introduction, or comment. All the annotation results are doubly checked.

\subsection{Heterogeneous Knowledge Expansion}

% Another important issue is to provide as sufficient knowledge that can be linked
% As discussed before, another consideration is
To provide sufficient candidate knowledge for each video, we conduct knowledge expansion by seeking information from two major relevant sources: 1) for the \textbf{semi- or unstructured knowledge}, we crawl the game-related data from an online soccer platform\textsuperscript{\ref{whoscores}}, including the \emph{player list}, \emph{player statistics}, \emph{team characteristics}, \emph{team formation} and the \emph{news about the game}. Note that we only preserve the data that is available before the game to prevent potential model crossing. 2) for the \textbf{structural knowledge}, we employ the annotated knowledge entities and game-related data to link each video to the Wikidata with BLINK~\cite{wu2020scalable} and crawl the $2$-hop related entity pages.
% Both two kinds of knowledge are integrated into a format that can be indexed by game IDs and entities.

\subsection{Data to Benchmark Adaption}

We further apply several treatments to make the data fit different video captioning methods.

% \textbf{Video Segmentation}: For video captioning tasks, the whole video of a game is kind of long (over $90$ minutes), so it is necessary to conduct video segmentation. We first follow and adapt the labeling of SoccerNet-v2 to complete the object detection (including multiple players, teams, and the ball) and event spotting (including $17$ types of soccer-specific events such as \emph{Goal}, \emph{Foul})~\cite{cioppa2021camera}. After that, we conduct the game video segmentation according to the key events and text length, producing over $8.3$k video clips with an average length of $7.97$ seconds.
\textbf{Video Segmentation}: As whole video of a game is kind of long (over $90$ minutes), so it is necessary to conduct video segmentation. We first follow and adapt the labeling of SoccerNet-v2 to complete the object detection (including multiple players, teams, and the ball) and event spotting (including $17$ types of soccer-specific events such as \emph{Goal}, \emph{Foul})~\cite{cioppa2021camera}. After that, we conduct the game video segmentation according to the key events and text length, producing over $8.9$k video clips with an average length of $10.31$ seconds.

\textbf{Feature Modeling}: As some of the existing methods require the pre-modeling of the 2D/3D visual features~\cite{zhang2020object,ye2022hierarchical}, we first utilize the player tracking toolkit~\footnote{\url{https://github.com/JooZef315/football-tracking-data-from-TV-broadcast}} to map the 3D video into 2D $1080*680$ football pitch, and then employ ResNet~\cite{he2016deep} to model both the 2D and 3D features as a prepared data source.

\subsection{Data Analysis}

% We present various characteristics of the GOAL benchmark by analyzing its statistics and comparing it with existing video captioning datasets.
We present various characteristics of GOAL and comparing it with existing benchmarks.

\textbf{Dataset Statistics.} Our dataset contains $8.9$k video clips, with an average of $2.46$ labeled sentences per video, for a total of $22$k sentences. Each video has an average of $21.67$ commentary words, which covers $81$\% of the entire video clip.

\begin{figure}
    \centering
    \includegraphics[width=1.0\linewidth]{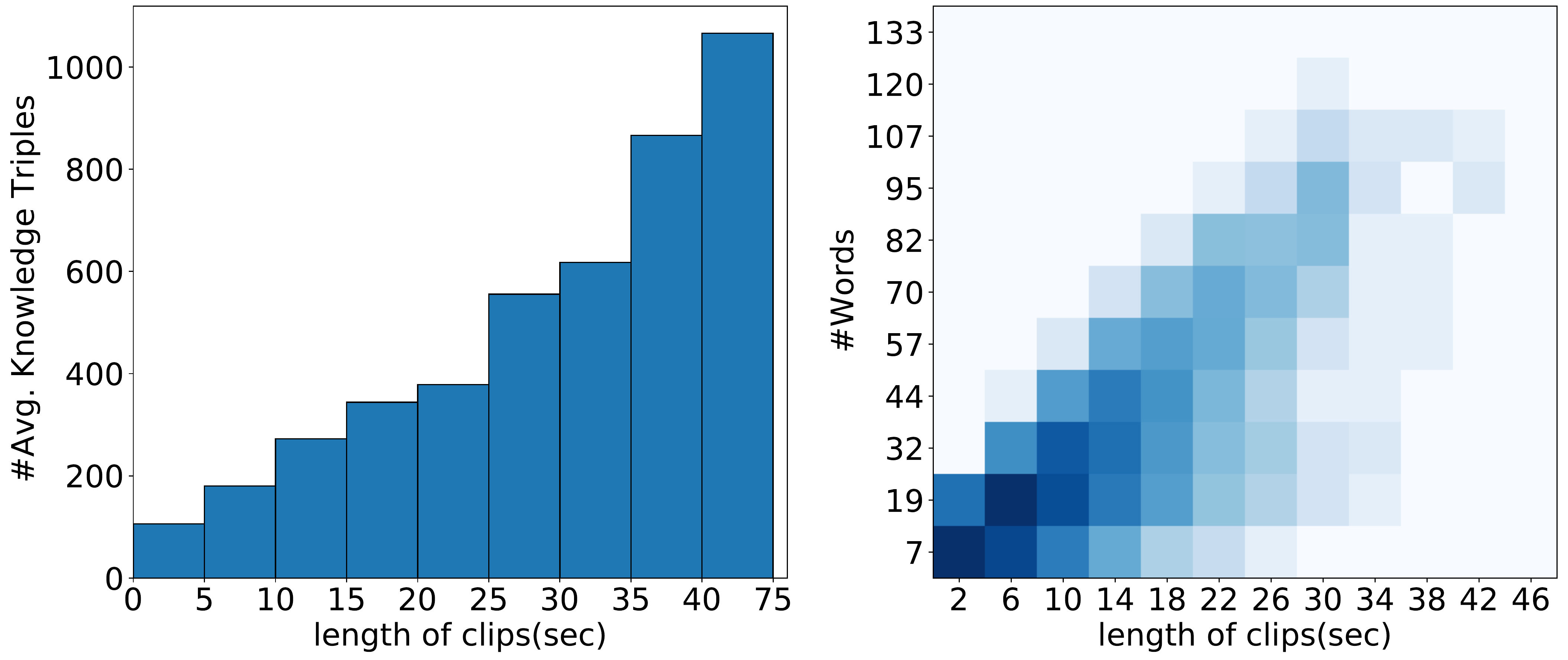}
    \caption{Distributions of the knowledge triples (left) and word number (right) over the video clip length.}
    \label{fig:distritbution}
\end{figure}

Meanwhile, each video is relevant to $193.1$ knowledge triples, and each video's text is labeled with an average of $1.84$ knowledge entities. Figure \ref{fig:distritbution} shows the related knowledge triples and word numbers both follow a nearly linear positive correlation with the clip length and keep stable fluctuations, which indicates that knowledge is evenly distributed in the sentences of the caption.

\textbf{Comparison with relevant datasets.}  The statistics in Table \ref{tab:dataset} further demonstrates that GOAL's setting is more challenging than the widely-used MSVD~\cite{chen2011collecting}, MSR-VTT~\cite{xu2016msr}, ActivityNet Captions~\cite{caba2015activitynet} and YouCook2~\cite{zhou2018towards}. We obtain their average similarity of video clips~\cite{yuan2020central}, and observe that GOAL's videos are the most similar, which indicates it requires more fine-grained recognizing and understanding abilities of the model (as most of the views are players running on a green field). Nevertheless, GOAL  contains the most words per second, the most diverse of the sentence (largest variance of sentence length), and pretty complex syntax (the depth of syntactic tree via~\citet{chen2014fast}), which makes it more challenging when generating texts.

\begin{table}[ht]
\small
\centering
\begin{tabular}{@{}lllll@{}}
\toprule
Dataset     & \begin{tabular}[c]{@{}l@{}}Video\\Simi.\end{tabular} & \begin{tabular}[c]{@{}l@{}} Words/ \\ Second\end{tabular} & \begin{tabular}[c]{@{}l@{}} Var. of\\Sent.$_{len}$ \end{tabular} & \begin{tabular}[c]{@{}l@{}}Syntactic\\ Complexity\end{tabular} \\ \midrule
MSVD        &   \underline{0.77}  &    \underline{0.78} & 2.86 &    6.38     \\
MSR-VTT     &   0.64  &    0.62 & 4.22 &  7.52     \\
ActivityNet &   0.61  &   0.22 & \underline{6.10} &   \textbf{8.64}   \\
YouCook2    &   0.58  &  0.19 &  4.58  &  6.43  \\ \hline
GOAL        &  \textbf{0.81}  &    \textbf{2.10} & \textbf{7.26} &       \underline{7.67}        \\ \bottomrule
\end{tabular}
\caption{Comparison of different datasets. Video Simi., Words/ second, Var.of Sent.$_{len}$ and Syntactic Complexity correspond to the average video similarity, words per second, variance of sentence length, and the average depth of syntactic tree of the captions. \textbf{Bolded} and \underline{underlined} represent the first and second largest value.}
\label{tab:dataset}
\end{table}

\section{Experiment}

We reproduce several representative video captioning methods on the GOAL benchmark and conduct an analysis of the experimental results.

\textbf{Setup.} We select the three types of methods: two-stage HMN~\cite{ye2022hierarchical}, end-to-end video-text pre-training SwinBERT~\cite{lin2022swinbert} and pre-training with prompting ALPRO~\cite{li2022align} as baselines, and additionally build knowledge features (for HMN and SwinBERT) or design knowledgeable prompts (for ALPRO) according to their model architectures as improvements. We adapt the common metrics (BLEU, METEOR~\cite{banerjee2005meteor}, Rouge-L and CIDEr~\cite{vedantam2015cider}) for evaluation.

\textbf{Major Result.} The experimental results are shown in Table \ref{tab:experiment}, from which we can observe that: (1) all three methods meet a severe decline on this dataset (e.g., SwinBERT's average \texttt{CIDEr} on MSR-VTT, YouCook2, MSVD, and ActivityNet is 95.5), indicating the challenging of this setting; (2) invoking knowledge is kind of benefit in improving existing methods, but designing appropriate prompts seems to be a more effective way (both HMN and SwinBERT are only slightly enhanced).

\begin{figure}
    \centering
    \includegraphics[width=1.0\linewidth]{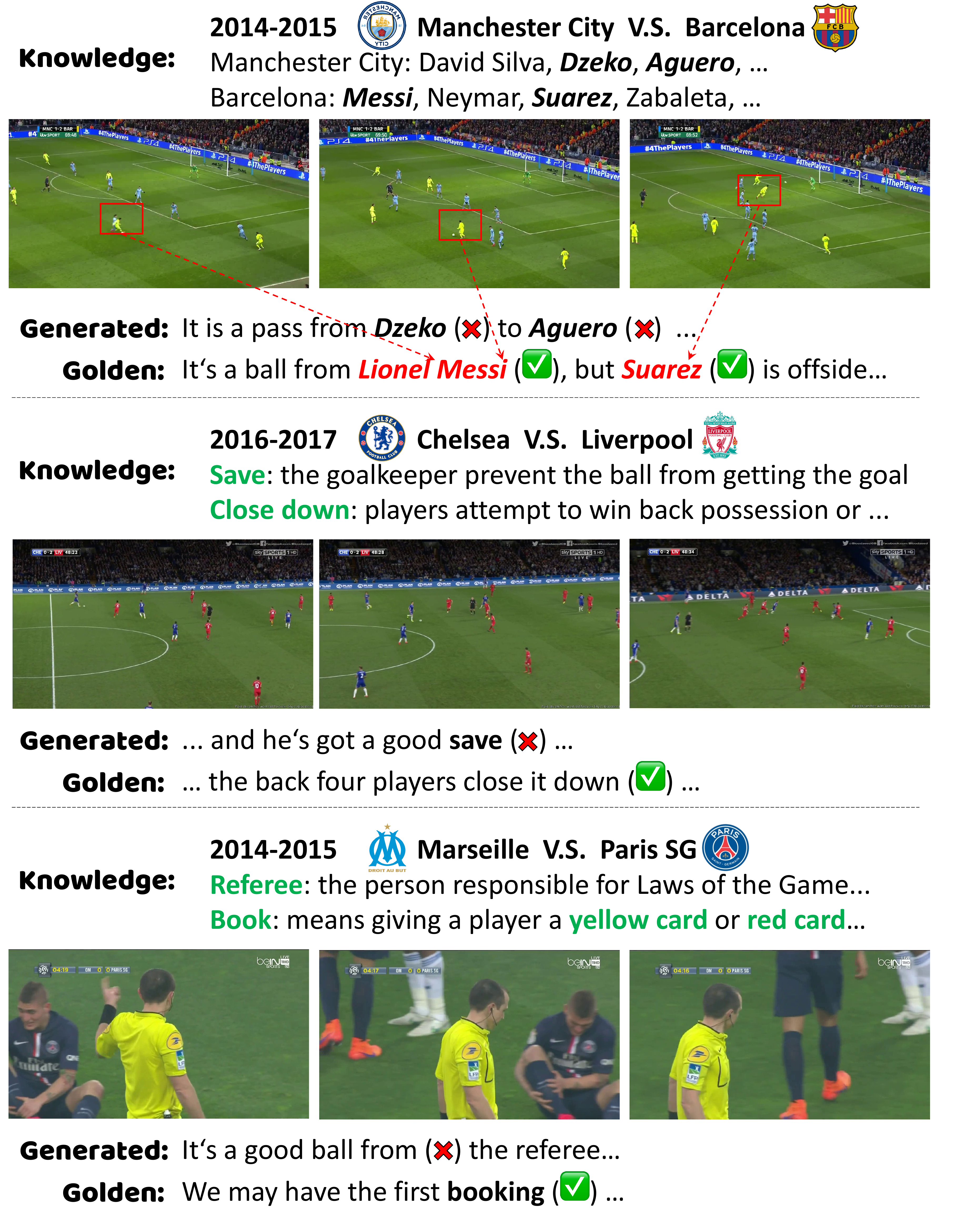}
    \caption{Representative error cases of the generated captions, which correspond to the entity mismatching, action misunderstanding and knowledge deficiency.}
    \label{fig:cases}
\end{figure}

\textbf{Error Analysis.} We conduct error analysis by presenting cases of ALPRO. Except for the general errors that models cannot effectively generate such a long and informative text, we notice several knowledge-aware errors, as shown in Figure \ref{fig:cases}. (1) Entity Mismatching: models often incorrectly recognize the fine-grained player objects, \emph{e.g.}, misidentify \emph{Messi} as \emph{Dzeko}, \emph{Suarez} as \emph{Aguero}. (2) Action Misunderstanding: models cannot understand a certain action with knowledge. Although \emph{Save} and \emph{Close down} are both actions in defense, they are totally different because the former one can only be performed by the goalkeeper. (3) Knowledge Deficiency: Current models lack a vast amount of parametric knowledge and the corresponding reasoning ability, which may cause many comical errors such as \emph{a good ball from the referee}.

\begin{table}[t]
\small
\centering
\resizebox{\linewidth}{!}{
\begin{tabular}{@{}lllll@{}}
\toprule
Method   & BLEU & METEOR & Rouge-L & CIDEr \\ \midrule
% SOTA$^{\texttt{MSVD}}$ &      &        &         &       \\
% SOTA$^{\texttt{MSR-VTT}}$ &      &        &         &       \\
% SOTA$^{\texttt{Acti.Net}}$ &      &        &         &       \\
% SOTA$^{\texttt{YouCook2}}$  &      &        &         &       \\ \hline
HMN &  13.5   &   5.1     &   11.1      &  3.3     \\
+Knowledge &  13.5    &    5.2    &    11.4↑     & 3.5↑      \\ \hline
SwinBERT &   10.3   &    5.6    &   11.1      &   3.7    \\
+Knowledge   &   10.4   &  5.5      &  12.0↑       &  4.4↑     \\  \hline
ALPRO    &   14.3   &    5.9    &   10.1      &   5.1    \\
+Knowledge   &  15.6↑     & 6.4↑       &   11.5↑      &  6.6↑     \\ \bottomrule
\end{tabular}}
\caption{Experimental results on the GOAL benchmark. We do not sign a ↑ if the value is only lifted within $0.1$.}
\label{tab:experiment}
\end{table}

\textbf{Discussion.} Given the above result, we discuss some potential directions for building advanced models for the KGVC task. First, beyond the current object detection, it is necessary to propose a knowledge-aware entity-object linking~\cite{wang2022wikidiverse} for such fine-grained video understanding applications. Second, KGVC models should exploit the language models' generation ability instead of just applying them in textual feature modeling. Third, considering the complexity of the task, it is promising to jointly utilize the chain-of-thought prompting of LLMs and knowledge graphs to understand, reason, and achieve better performance.

\section{Conclusion and Future Work}

In this paper, we propose the task of knowledge-grounded video captioning (KGVC) and present a challenging benchmark upon real-time soccer commentary, GOAL, for supporting the research explorations. We conduct preliminary experiments to prove the difficulty of our benchmark and propose several promising directions.
Emergent future work includes: 1) designing fine-grained knowledgeable captioning methods as well as knowledge-aware evaluation metrics; 2) exploiting the large-scale language models for reasoning in this task; 3) expanding the knowledge-grounded video captioning setting to other real-world applications.

% \newpage

\section{Limitations}
As a pioneering exploration of knowledge-grounded video captioning, our work still preserves several limitations that need to be improved by ourselves and other interested researchers.

For the proposed dataset, we summarize three main limitations. First, due to the difficulties of collecting open-source soccer videos, each video only has one version of the commentary caption, making it too strict for evaluating models. Second, although we collect most of the relevant knowledge that can be linked to the video or the text, we still ignore some other types of knowledge, such as the soccer commentator's style. Third, the scale of the current dataset is not large enough, which is emergent to be enriched in our subsequent research.

For the experiments, we only select three representative top-performing methods due to diverse reasons, such as the workload of adapting to our dataset, loss of open-source code, etc. Meanwhile, beyond our primary attempts, there are many potential ways to improve current models with external knowledge, which should be further explored.

Despite these limitations, we believe our idea of building a knowledge-aware, applicable setting of video captioning task is valuable, and our proposed GOAL benchmark can be a valuable data source to serve the pioneer researchers with its videos, high-quality captions and knowledge sources.

\section{Ethic Consideration}

There are two major concerns when constructing our new benchmark: (1) Video Copyright: all the videos and speeches are downloaded from the open-source SoccerNet-v2~\cite{Deliege2020SoccerNetv2}, which is under the MIT license. All the data collection does not involve commercial activities. (2) Annotator Right: all the annotators in our team are paid a salary higher than the market average. We allow each annotator to choose the working time and the exit timing freely. After the overall data quality verification, we additionally offer gifts to them as a souvenir for this memorable project.

\bibliography{acl2023}
\bibliographystyle{acl_natbib}

\appendix

\section{Appendix}
\label{sec:appendix}

\subsection{Background and Related Work}

% \section{Related Work}

As an essential task that connects the video and texts, video captioning has a wide range of real-world applications such as human-robot interaction, sports narrative, and describing videos for the blind~\cite{sechet1980antiope}. To support the relevant technical explorations, the early researchers first construct open datasets by collecting general videos from websites such as YouTube~\cite{chen2011collecting,xu2016msr}, which swiftly attracts a series of efforts to build neural networks for solving this task~\cite{he2016deep,gao2017video}. Along with the growing model performance, researchers gradually pivot to proposing and tackling more difficult task settings, such as captioning with commonsense~\cite{fang2020video2commonsense}, dense event~\cite{zhou2018towards} and fine-grained actions~\cite{yu2018fine}. However, some of the challenging video captioning datasets are actually not publicly available~\cite{jain2022video}.
While finishing this study, a concurrent work~\citep{suglia2022going} also named GOAL was released, which annotated 1,107 football video highlights from YouTube. However, we build our dataset based on the complete match videos from SoccerNet to be compatible with existing official resources, which makes the distinct focuses.

Although this task requires several abilities of the model, current video captioning methods mainly focus on the \emph{video understanding}~\cite{xu2021vlm,shi2022learning} and \emph{video-text bridging}~\cite{ge2022bridging} because the existing datasets are commonly consists of raw, short descriptions for coarse-grained scenes. However, we argue that \emph{text generation}~\cite{lin2021vx2text} deserves more attention towards building a real-world captioning application. With the emergence of large-scale pre-trained models~\cite{brown2020language,ouyang2022training}, it is promising to invoke abundant external knowledge, conduct complex reasoning, and generation to explore more challenging settings of this task.

% \subsection{Annotation Process}

\subsection{Annotation Process}

Before joining the annotation group, each person is informed about the utilization of the annotation results. After the annotation starts, we randomly send the preserved $20$ videos to the $10$ annotators in our team while each annotator is asked to finish $4$ videos so that each video is guaranteed to be labeled by two people. For each task, we take an extra 10-minute game as a demonstration to explain the details before the large-scale annotation. Once there is a disagreement, we ask another annotator to decide the more appropriate one anonymously. Specifically, for the tough job of \textbf{video-text alignment}, we conduct a collective review by watching the videos (the version with captions) together and filtering out the mistakes during this process.

\begin{figure*}[ht]
    \centering
    \includegraphics[width=1.0\linewidth]{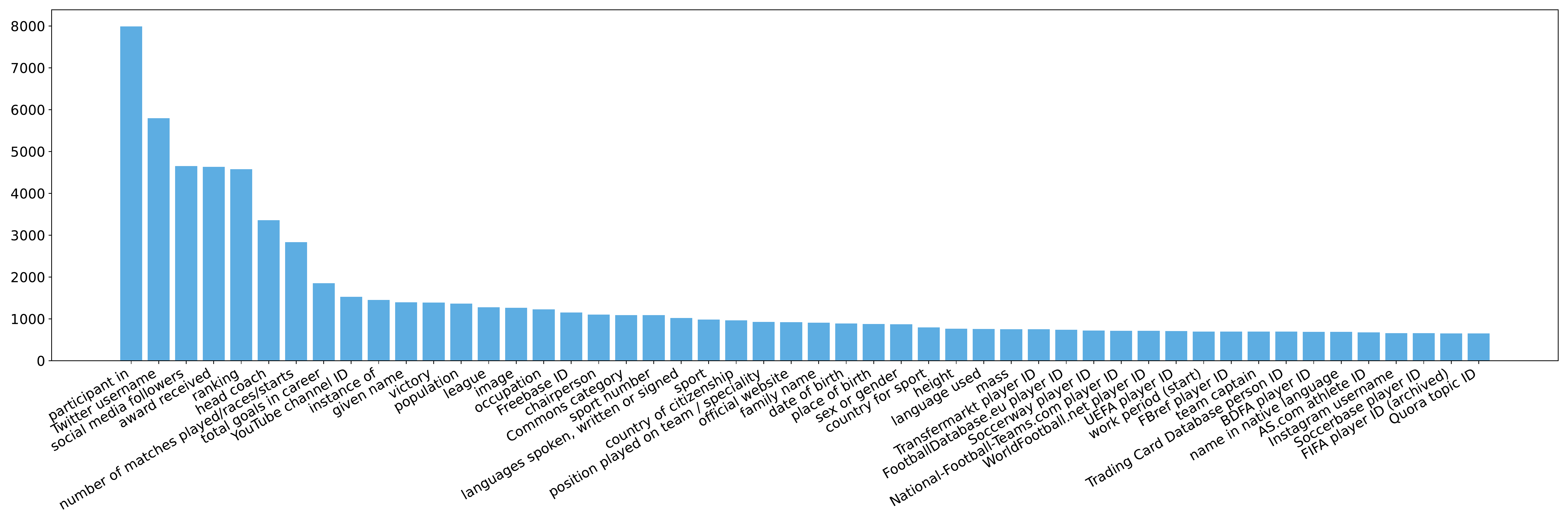}
    \caption{The distribution of the top-$50$ relations in our dataset.}
    \label{fig:relation}
\end{figure*}

\subsection{Implementation Details}

All the models are trained with a server with $8$ Nvidia $3090$ GPUs, each with $24$G VRAM, $24$ AMD CPU cores, with $251$GB RAM. When adapting baselines, all hyper-parameters, including learning-rate and mask probability, etc., are tuned.

For the knowledge modeling for HMN and SwinBERT, we mainly select the entities about players and teams. We design rules to transfer them into a paragraph and employ BERT~\cite{devlin2019bert} to obtain the feature.
Meanwhile, as the ALPRO is implemented as a video question-answering model, we only preserve the encoder of it and utilize T5~\cite{raffel2020exploring} as a decoder to complete the generation. To bridge the two modules, we employ a simple prompt ``\texttt{What is the commentary on the current video clip?}'' to conduct the generation. When turning to the knowledgeable version, such a prompt is combined with the previous knowledge paragraph for HMN and SwinBERT.

\begin{figure}
    \centering
    \includegraphics[width=1.0\linewidth]{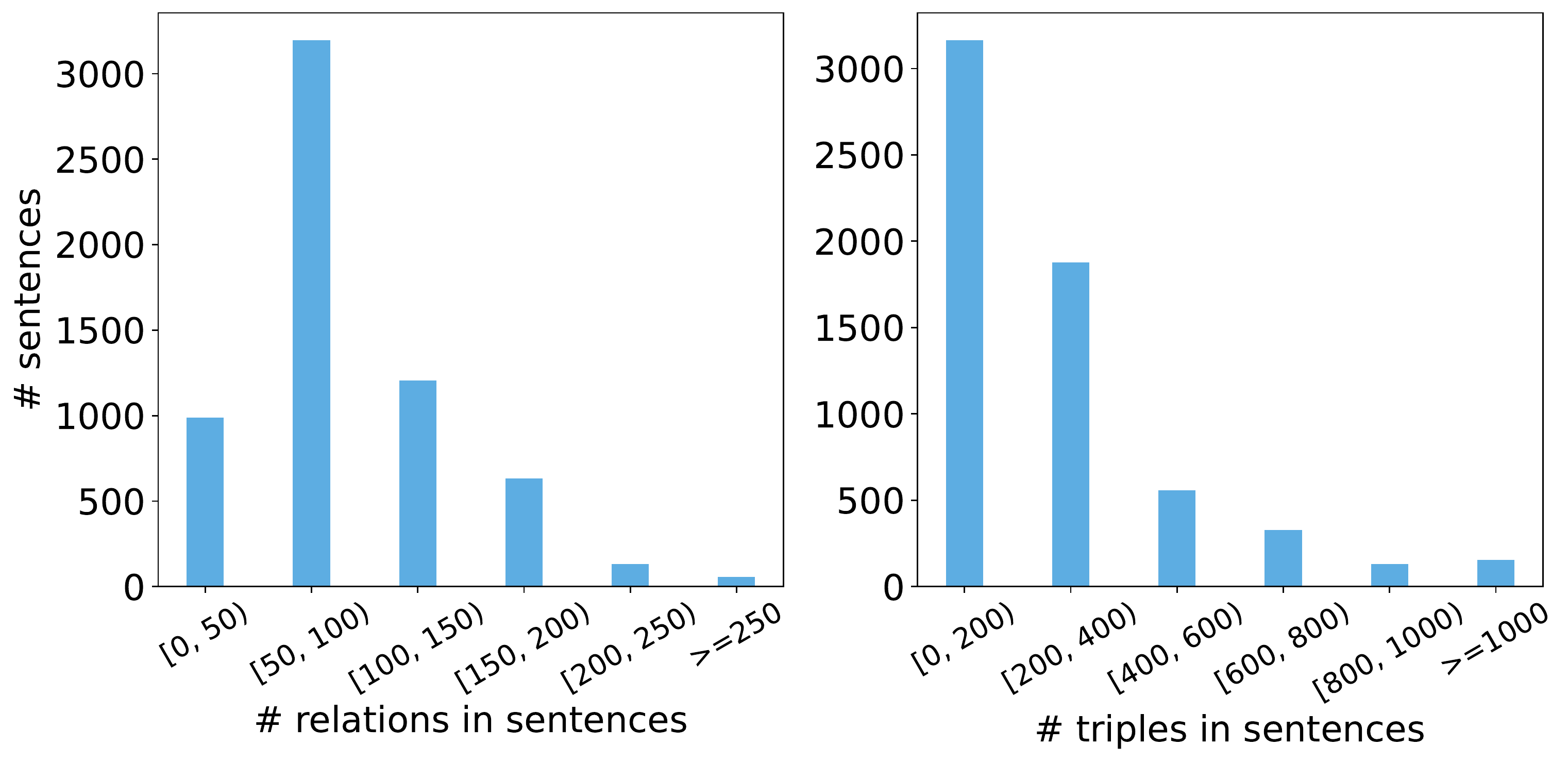}
    \caption{The distribution of sentence-level knowledge density in terms of relation types and knowledge triples.}
    \label{fig:sentence-knowledge}
\end{figure}

\begin{figure}
    \centering
    \includegraphics[width=1.0\linewidth]{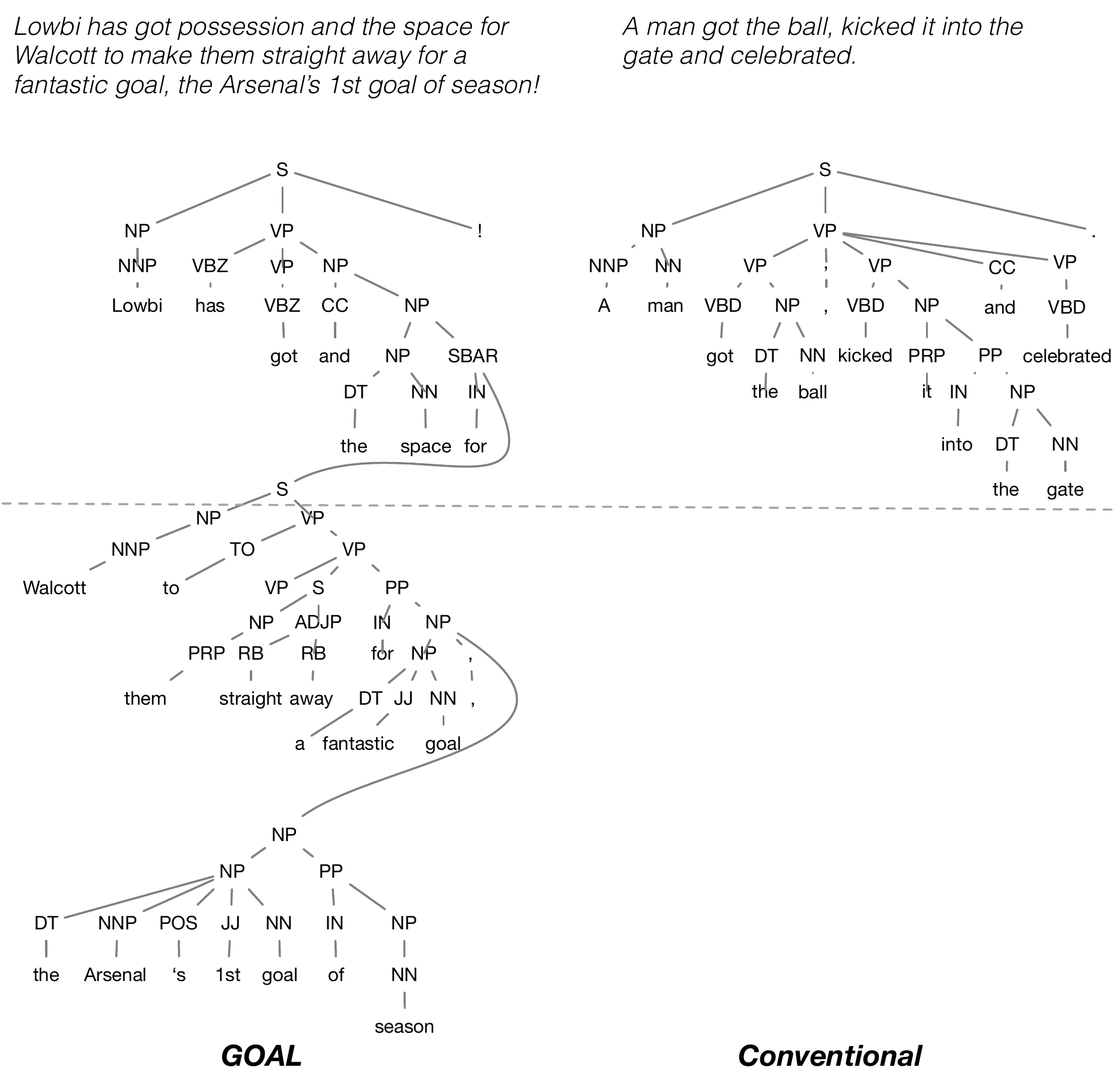}
    \caption{The syntax-level comparison of the case in GOAL and conventional video captioning dataset.}
    \label{fig:syntax}
\end{figure}

\subsection{Additional Data Analysis}

Except for the analysis introduced in the main paper, we conduct a series of studies to perceive the knowledge distribution in the GOAL benchmark.

\textbf{Knowledge Distribution}. Figure \ref{fig:relation} shows the distribution of the top-$50$ relations in the GOAL dataset, from which we can infer that most of the knowledge is about players. Except for the personalized properties, the historical background is also an important part of the knowledge. Meanwhile, the sentences in our dataset are mostly associated with abundant knowledge, whether in terms of relation types and knowledge triples, as shown in Figure \ref{fig:sentence-knowledge}. This further confirms the richness of knowledge of our dataset.

\textbf{A Case of Syntax Analysis}. For the example shown in Section \ref{sec:introduction}, we present a syntax-level analysis case to present the hardness of the GOAL benchmark. The result shown in Figure \ref{fig:syntax} is produced by the Stanford Syntax Parser~\cite{chen2014fast}. The syntactic tree of our dataset is deeper than others, which indicates that GOAL is more challenging to be solved.

\end{document}